\RequirePackage[hyphens]{url}
\documentclass{article}

\usepackage{microtype}
\usepackage{graphicx}
\usepackage{subfigure}
\usepackage{booktabs} 

\usepackage{hyperref}

\usepackage[accepted]{icml2024}

\usepackage{amsmath}
\usepackage{amssymb}
\usepackage{mathtools}
\usepackage{amsthm}

\usepackage[capitalize,noabbrev]{cleveref}

\theoremstyle{plain}
\newtheorem{theorem}{Theorem}[section]
\newtheorem{proposition}[theorem]{Proposition}

\theoremstyle{definition}

\theoremstyle{remark}

\usepackage[textsize=tiny]{todonotes}

\icmltitlerunning{Position: Stop Making Unscientific AGI Performance Claims}

\begin{document}

\twocolumn[
\icmltitle{Position: Stop Making Unscientific AGI Performance Claims}



\icmlsetsymbol{equal}{*}

\begin{icmlauthorlist}
\icmlauthor{Patrick Altmeyer}{equal,yyy}
\icmlauthor{Andrew M. Demetriou}{equal,yyy}
\icmlauthor{Antony Bartlett}{yyy}
\icmlauthor{Cynthia C. S. Liem}{yyy}
\end{icmlauthorlist}

\icmlaffiliation{yyy}{Department of Intelligent Systems, Delft University of Technology, Delft, the Netherlands}

\icmlcorrespondingauthor{Patrick Altmeyer}{patrick.altmeyer@tudelft.nl}

\icmlkeywords{Machine Learning, Anthropomorphism, Artificial General Intelligence}

\vskip 0.3in
]



\printAffiliationsAndNotice{\icmlEqualContribution} 

\begin{abstract}
Developments in the field of Artificial Intelligence (AI), and particularly large language models (LLMs), have created a `perfect storm’ for observing `sparks’ of Artificial General Intelligence (AGI) that are spurious. Like simpler models, LLMs distill meaningful representations in their latent embeddings that have been shown to correlate with external variables. Nonetheless, the correlation of such representations has often been linked to human-like intelligence in the latter but not the former. We probe models of varying complexity including random projections, matrix decompositions, deep autoencoders and transformers: all of them successfully distill information that can be used to predict latent or external variables and yet none of them have previously been linked to AGI. We argue and empirically demonstrate that the finding of meaningful patterns in latent spaces of models cannot be seen as evidence in favor of AGI. Additionally, we review literature from the social sciences that shows that humans are prone to seek such patterns and anthropomorphize. We conclude that both the methodological setup and common public image of AI are ideal for the misinterpretation that correlations between model representations and some variables of interest are `caused' by the model's understanding of underlying `ground truth’ relationships. We, therefore, call for the academic community to exercise extra caution, and to be keenly aware of principles of academic integrity, in interpreting and communicating about AI research outcomes.
\end{abstract}

\section{Introduction}\label{intro}

In 1942, when anti-intellectualism was rising and the integrity of science was under attack, Robert K.\ Merton formulated four `institutional imperatives' as comprising the ethos of modern science: \emph{universalism}, meaning that the acceptance or rejection of claims entering the lists of science should not depend on personal or social attributes of the person bringing in these claims; \emph{``communism''} [sic], meaning that there should be common ownership of scientific findings and one should communicate findings, rather than keeping them secret; \emph{disinterestedness}, meaning that scientific integrity is upheld by not having self-interested motivations, and \emph{organized skepticism}, meaning that judgment on the scientific contribution should be suspended until detached scrutiny is performed, according to institutionally accepted criteria~\citep{merton1942science}. While the Mertonian norms may not formally be known to academics today, they still are implicitly being subscribed to in many ways in which academia has organized academic scrutiny; e.g., through the adoption of double-blind peer reviewing, and in motivations behind open science reforms.

At the same time, in the way in which academic research is disseminated in the AI and machine learning fields today, major shifts are happening. Where these research fields have actively adopted early sharing of preprints and code, the volume of publishable work has exploded to a degree that one cannot reasonably keep up with broad state-of-the-art, and social media influencers start playing a role in article discovery and citeability~\citep{weissburg2024tweets}. Furthermore, because of major commercial stakes with regard to AI and machine learning technology, and e.g.\ following the enthusiastic societal uptake of products employing LLMs, such as ChatGPT, the pressure to beat competitors as fast as possible is only increasing, and strong eagerness can be observed in many domains to `do something with AI' in order to innovate and remain current.

Where AI used to be a computational modeling tool to better understand human cognition~\citep{vanrooij2023aitheoretical}, the recent interest in AI and LLMs has been turning into one in which AI is seen as a tool that can mimic, surpass and potentially replace human intelligence. In this, the achievement of Artificial General Intelligence (AGI) has become a grand challenge, and in some cases, an explicit business goal. The definition of AGI itself is not as clear-cut or consistent; loosely, it is a phenomenon contrasting with `narrow AI' systems, that were trained for specific tasks~\citep{goertzel2014artificial}. In practice, to demonstrate that the achievement of AGI may be getting closer, researchers have sought to show that AI models generalize to different (and possibly unseen) tasks, with little human intervention, or show performance considered `surprising' to humans.

For example, Google DeepMind claimed their AlphaGeometry model~\citep{trinh2024geometry} reached a `milestone' towards AGI. This model has the ability to solve complex geometry problems, allegedly without the need for human demonstrations during training. However, work such as this had been initially introduced in the 1950s~\citep{hector2024linkedin}: without the use of an LLM, logical inference systems proved 100\% accurate in proving all the theorems of Euclidean Geometry, due to geometry being an axiomatically closed system. Therefore, while DeepMind created a powerfully fast geometry-solving machine, it is still far from AGI.

Generally, in the popularity of ChatGPT and the integration of generative AI in productivity tools (e.g.\ through Microsoft's Copilot integrations in GitHub and Office applications), one also can wonder whether the promise of AI is more in computationally achieving general intelligence, or rather in the engineering of general-purpose tools\footnote{A Swiss army knife is an effective general-purpose tool, without people wondering whether it exhibits intelligence.}. Regardless, stakes and interests are high, e.g.\ with ChatGPT clearing nearly \$1 billion in months of its release\footnote{\url{https://www.bloomberg.com/news/articles/2023-08-30/openai-nears-1-billion-of-annual-sales-as-chatgpt-takes-off}}.

When combining massive financial incentives with the presence of a challenging and difficult-to-understand technology, that aims towards human-like problem-solving and communication abilities, a situation arises that is fertile for the misinterpretation of spurious cues as hints towards AGI, or other qualities like sentience\footnote{\url{https://www.scientificamerican.com/article/google-engineer-claims-ai-chatbot-is-sentient-why-that-matters/}} and consciousness. AI technology only becomes more difficult to understand as academic publishing in the space largely favors performance, generalization, quantitative evidence, efficiency, building on past work, and novelty~\citep{values_in_ML}. As such, works that make it into top-tier venues tend to propose heavier and more complicated technical takes on tasks that (in the push towards generalizability) get more vague, while the scaling-up of data makes traceability of possible memorization harder. In a submission-overloaded reality, researchers may further get incentivized to oversell and overstate achievement claims. At the same time, while currently popular in literature, inherent complexity and opaqueness in technical solutions may fundamentally be unwise to pursue in high-stakes applications~\citep{rudin2019stop}.

Noticing these trends, we as the authors of this article are concerned. We feel that the current culture of racing toward Big Outcome Statements in industry and academic publishing too much disincentivizes efforts toward more thorough and nuanced actual problem understanding. At the same time, as the outside world is so eager to adopt AI technology, (too) strong claims make for good sales pitches, but a question is whether there is indeed sufficient evidence for these claims. With successful AGI outcomes needing to look human-like, this also directly plays into risks of anthropomorphizing (the attribution of human-like qualities to non-human objects) and confirmation bias (the seeking-out and/or biased interpretation of evidence in support of one's beliefs). In other words, it is very tempting to claim surprising human-like achievements of AI, and as humans, we are very prone to genuinely believing this. \textbf{We therefore urge our fellow researchers to stop making unscientific AGI performance claims}.

To strengthen our argument, in this paper, we first present related work in Section~\ref*{related}. We then consider a recently viral work~\citep{gurnee2023languagev2} in which claims about the learning of world models by LLMs were made. In Section~\ref*{patterns-in-latent-spaces-and-how-to-find-them}, we present several experiments that may invite similar claims on models yielding more intelligent outcomes than would have been expected---while at the same time indicating how we feel these claims should \emph{not} be made. Furthermore, we present a review of social science findings in Section~\ref*{social} that underline how prone humans are to being enticed by patterns that are not really there. Combining this with the way in which media portrayal of AI has tended towards science-fiction imagery of mankind-threatening robots, we argue that the current AI culture is a perfect storm for making and believing inflated claims, and call upon our fellow academics to be extra mindful and scrutinous about this. Finally, in Section~\ref*{outlook}, we propose specific structural and cultural changes to improve the current situation. Section~\ref*{conclude} concludes. 

\section{Related Work}\label{related}

In this work, we question the practice of using outcomes from mechanistic interpretability to support AGI claims. This is not to be seen as criticism toward the underlying methodologies in isolation, but rather in the context of current publishing practices that we repeatedly challenge throughout this work. Many closely related works are free of any grandiose conclusions and instead highlight the benefits of mechanistic interpretability that we also highlight here~\citep{nanda2023emergent,gurnee2023finding,li2022emergent}. 

Another related subfield investigates the capacity of LLMs to reason causally. Here, too, there is an opportunity to over-interpret the finding of causal information as causal understanding. Recent work has shown, for example, that LLMs can indeed correctly predict causal relationships and this may have practical use cases~\citep{kiciman2023causal}. But despite the potential utility, the authors also demonstrate that this capacity can be partially explained by memorization, rather than an actual understanding of causal relationships. Similarly,~\citet{zevcevic2023causal} provide evidence indicating that current LLMs ``may talk causality but are not causal''.

Two other recent works are related to this work and align well with the position we present here.~\citet{schaeffer2024emergent} demonstrate that the apparent emergent abilities of large language models may be driven by a choice of evaluation metrics, rather than some fundamental property that is intrinsic to this family of models. Their work highlights the need for rigorous testing and benchmarking of LLMs, which we also point to in this work, albeit in a slightly different methodological context.~\citet{kloft2024ai} provide experimental evidence demonstrating that people have heightened expectations and a biased, positive view of AI. The authors run a user study of human-AI interaction, in which participants performed better at a given task when they (wrongly) thought they were aided by a positively described AI. This placebo effect was found to be robust to negative descriptions of AI. Positive bias towards AI may exacerbate other factors that drive people to make unscientific claims about the current state of AI, which we discuss in Section~\ref*{social}.

\section{Surprising Patterns in Latent Spaces?}\label{patterns-in-latent-spaces-and-how-to-find-them}

In 2023, a research article went viral on the X\footnote{\url{https://twitter.com/wesg52/status/1709551516577902782?s=20}} platform~\citep{gurnee2023languagev1}. Through linear probing experiments, the claim was made that LLMs learned literal maps of the world. As such, they were considered to be more than `stochastic parrots'~\citep{bender2021dangers} that can only correlate and mimic existing patterns from data, but not truly understand it. While the manuscript immediately received public criticism~\citep{marcus2023muddles}, and the revised, current version is more careful with regard to its claims~\citep{gurnee2023languagev2}, reactions on X seemed to largely exhibit excitement and surprise at the authors' findings. However, in this section, through various simple examples, we make the point that observing patterns in latent spaces should not be a surprising revelation. After starting with a playful example of how easy it is to `observe' a world model, we build up a larger example focusing on key economic indicators and central bank communications.

\begin{figure}

{\centering \includegraphics[width=0.5\textwidth]{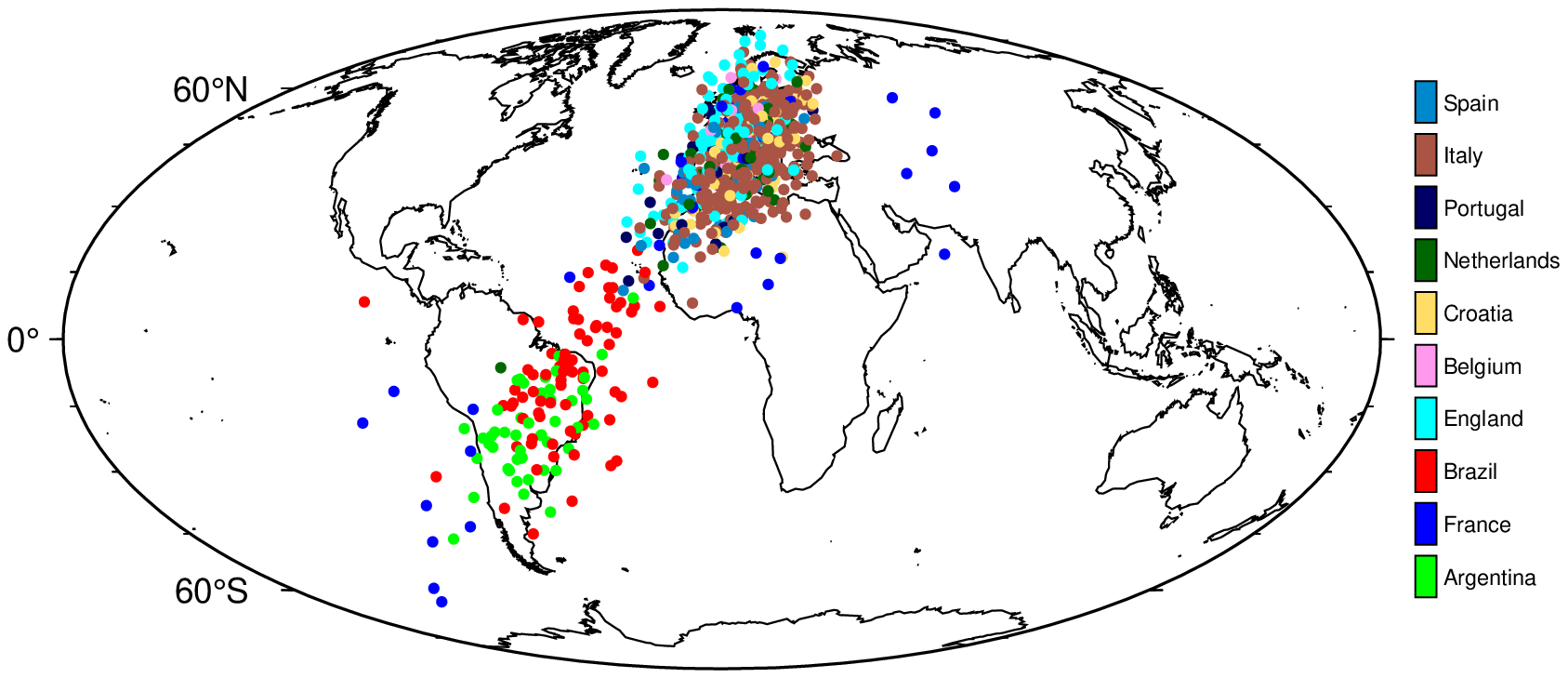}

}

\caption{\label{fig-map}Predicted coordinate values (out-of-sample) from a linear probe on final-layer activations of an untrained neural network.}

\end{figure}

\subsection{Are Neural Networks Born with World Models?}\label{example-deep-learning}

\citet{gurnee2023languagev2} extract and visualize the alleged geographical world model by training linear regression probes on internal activations in LLMs (including Llama-2) for the names of places, to predict geographical coordinates associated with these places. Now, the Llama-2 model has ingested huge amounts of publicly available data from the internet, including Wikipedia dumps from the June-August 2022 period \citep{touvron2023llama}. It is therefore highly likely that the training data contains geographical coordinates, either directly or indirectly. At the very least, we should expect that the model has seen features during training that are highly correlated with geographical coordinates. The model itself is essentially a very large latent space to which all features are randomly projected in the very first instance before being passed through a series of layers which are gradually trained for downstream tasks.

In our first example, we simulate this scenario, stopping short of training the model. In particular, we take the \href{https://github.com/wesg52/world-models/blob/main/data/entity_datasets/world_place.csv}{world\_place.csv} that was used in \citet{gurnee2023languagev2}, which maps locations/areas to their latitude and longitude. For each place, it also indicates the corresponding country. From this, we take the subset that contains countries that are currently part of the top 10 \href{https://www.fifa.com/fifa-world-ranking/men?dateId=id14142}{FIFA world ranking}, and assign the current rank to each country (i.e., Argentina gets 1, France gets 2, \ldots{}). To ensure that the training data only involves a noisy version of the coordinates, we transform the longitude and latitude data as follows: \(\rho \cdot \text{coord} + (1-\rho) \cdot \epsilon\) where \(\rho=0.5\) and \(\epsilon \sim \mathcal{N}(0, 5)\).

Next, we encode all features except the FIFA world rank indicator as continuous variables: \(X^{(n \times m)}\) where \(n\) is the number of samples and \(m\) is the number of resulting features. Additionally, we add a large number of random features to \(X\) to simulate the fact that not all features ingested by Llama-2 are necessarily correlated with geographical coordinates. Let \(d\) denote the final number of features, i.e.~\(d=m+k\) where \(k\) is the number of random features.

We then initialize a small neural network, considered a \emph{projector}, mapping from \(X\) to a single hidden layer with \(h<d\) hidden units and sigmoid activation, and from there, to a lower-dimensional output space. Without performing any training on the \emph{projector}, we simply compute a forward pass of \(X\) and retrieve activations \(\mathbf{Z}^{(n\times h)}\). Next, we perform the linear probe on a subset of \(\mathbf{Z}\) through Ridge regression: \(\mathbf{W} = (\mathbf{Z}_{\text{train}}'\mathbf{Z}_{\text{train}} + \lambda \mathbf{I}) (\mathbf{Z}_{\text{train}}'\textbf{coord})^{-1}\), where \(\textbf{coord}\) is the \((n \times 2)\) matrix containing the longitude and latitude for each sample. A hold-out set is reserved for testing, on which we compute predicted coordinates for each sample as \(\widehat{\textbf{coord}}=\mathbf{Z}_{\text{test}}\mathbf{W}\) and plot these on a world map (Figure~\ref{fig-map}). 

While the fit certainly is not perfect, the results do indicate that the random projection contains representations that are useful for the task at hand. Thus, this simple example illustrates that meaningful target representations should be recoverable from a sufficiently large latent space, given the projection of a small number of highly correlated features. Similarly, \citet{alain2018understanding} observe that even before training a convolutional neural network on MNIST data, the layer-wise activations can already be used to perform binary classification. In fact, it is well-known that random projections can be used for prediction tasks \citep{dasgupta2013experiments}.

This first experiment---and indeed the practice of probing LLMs that have seen vast amounts of data---can be seen as a form of inverse problem and common caveats such as non-uniqueness and instability apply~\citep{haltmeier2023regularization}. Regularization can help mitigate these caveats~\citep{om2001ridge}, but we confess that we did not carefully consider the parameter choice for $\lambda $, nor has this been carefully studied in the related literature to the best of our knowledge.

\subsection{PCA as a Yield Curve Interpreter}\label{example-principal-component-analysis}

We now move to a concrete application domain: Economics. Here, the yield curve, plotting the yields of bonds against their maturities, is a popular tool for investors and economists to gauge the health of the economy. The yield curve's slope is often used as a predictor of future economic activity: a steep yield curve is associated with a growing economy, while a flat or inverted yield curve is associated with a contracting economy. To leverage this information in downstream modelling tasks, economists regularly use PCA to extract a low-dimensional projection of the yield curve that captures relevant variation in the data (e.g.\ \citet{berardi2022dissecting}, \citet{kumar2022effective} and \citet{crump2019deconstructing}).

To understand the nature of this low-dimensional projection, we collect daily Treasury par yield curve rates at all available maturities from the US Department of the Treasury. Computing principal components involves decomposing the matrix of all yields \(\mathbf{r}\) into a product of its singular vectors and values: \(\mathbf{r}=\mathbf{U}\Sigma\mathbf{V}^{\prime}\). Let us simply refer to \(\mathbf{U}\), \(\Sigma\) and \(\mathbf{V}^{\prime}\) as latent embeddings of the yield curve.

\begin{figure*}[tb]

\centering{

\includegraphics[width=0.9\textwidth]{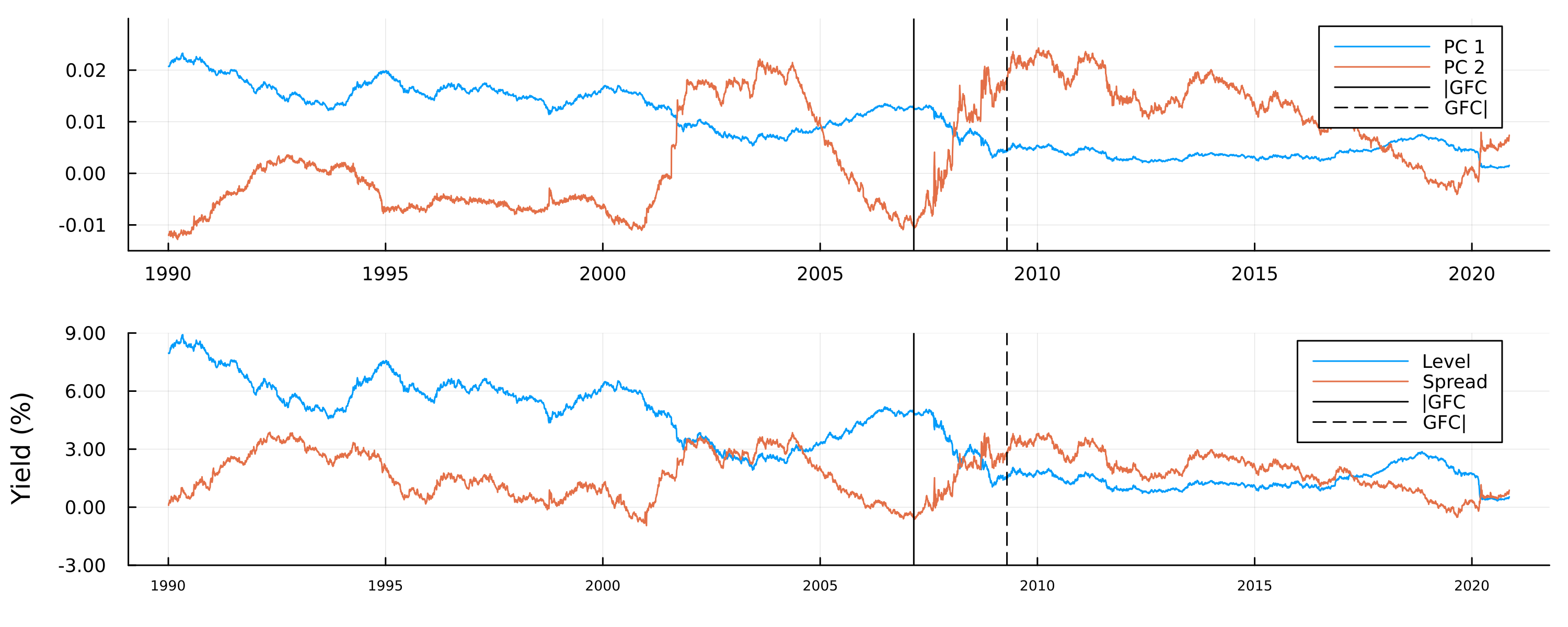}

}

\caption{\label{fig-pca-yield}Top chart: The first two principal components of US Treasury yields over time at daily frequency. Bottom chart: Observed average level and 10yr-3mo spread of the yield curve. Vertical stalks roughly indicate the onset ($|$GFC) and the beginning of the aftermath (GFC$|$) of the Global Financial Crisis.}

\end{figure*}

The upper panel in Figure~\ref{fig-pca-yield} shows the first two principal components of the yield curves of US Treasury bonds over time. Vertical stalks indicate key dates related to the Global Financial Crisis (GFC). During its onset, on 27 February 2007, financial markets were in turmoil following a warning from the Federal Reserve (Fed) that the US economy was at risk of a recession. The Fed later reacted to mounting economic pressures by gradually reducing short-term interest rates to unprecedented lows. Consequently, the average level of yields decreased and the curve steepened. In Figure~\ref{fig-pca-yield}, we can observe that the first two principal components appear to capture this level shift and steepening, respectively. In fact, they are strongly positively correlated with the actual observed first two moments of the yield curve (lower panel of Figure~\ref{fig-pca-yield}).

Again, it should not be surprising that these latent embeddings are meaningful: by construction, principal components are orthogonal linear combinations of the data itself, each of which explains most of the residual variance after controlling for the effect of all previous components.

\subsection{LLMs for Economic Sentiment Prediction}\label{ex-llm}

So far, we considered simple linear data transformations. One might argue that this does not really involve latent embeddings in the way they are typically thought of in the context of deep learning. In Appendix~\ref{appendix:autoencoder}, we present an additional experiment in which we more explicitly seek neural network-based representations that will be useful for downstream tasks. Here, we continue with an example in which LLMs may be used for economic sentiment prediction.

Closely following the approach in \citet{gurnee2023languagev2}, we apply it to the novel \emph{Trillion Dollar Words}~\citep{shah2023trillion} financial dataset, containing a curated selection of sentences formulated and communicated to the public by the Fed through speeches, meeting minutes and press conferences.~\citep{shah2023trillion} use this dataset to train a set of LLMs and rule-based models to classify sentences as either `dovish', `hawkish' or `neutral'. In the context of central banking, `hawkishness' is typically associated with tight monetary policy: in other words, a `hawkish' stance on policy favors high interest rates to limit the supply of money and thereby control inflation. The authors first manually annotate a sub-sample of the available data and then fine-tune various models for the classification task. Their model of choice, \emph{FOMC-RoBERTa} (a fine-tuned version of RoBERTa \citep{liu2019roberta}), achieves an \(F_1\) score of around \(>0.7\) on the test data. To illustrate the potential usefulness of the learned classifier, they use predicted labels for the entire dataset to compute an ad-hoc, count-based measure of `hawkishness'. This measure is shown to correlate with key economic indicators in the expected direction: when inflationary pressures rise, the measured level of `hawkishness' increases, as central bankers react by raising interest rates to bring inflation back to target.

\subsubsection{Linear Probes}\label{linear-probes}
We now use linear probes to assess if the fine-tuned model has learned associative patterns between central bank communications and key economic indicators. Therefore, we further pre-process the data provided by \citet{shah2023trillion} and use their proposed model to compute activations of the hidden state, on the first entity token for each layer. We have made these available and easily accessible through a small Julia package: \href{https://anonymous.4open.science/r/TrillionDollarWords/README.md}{TrillionDollarWords.jl}. 

For each layer, we compute linear probes through Ridge regression on two inflation indicators (the Consumer Price Index (CPI) and the Producer Price Index (PPI)) and US Treasury yields at different levels of maturity. To allow comparison with \citet{shah2023trillion}, we let yields enter the regressions in levels. To measure price inflation we use percentage changes proxied by log differences. To mitigate issues related to over-parameterization, we follow the recommendation in \citet{alain2018understanding} to first reduce the dimensionality of the computed activations. In particular, we restrict our linear probes to the first 128 principal components of the embeddings of each layer. To account for stochasticity, we use an expanding window scheme with 5 folds for each indicator and layer. To avoid look-ahead bias, PCA is always computed on the sub-samples used for training the probe. 

\begin{figure*}

\centering{

\includegraphics[width=1.0\textwidth]{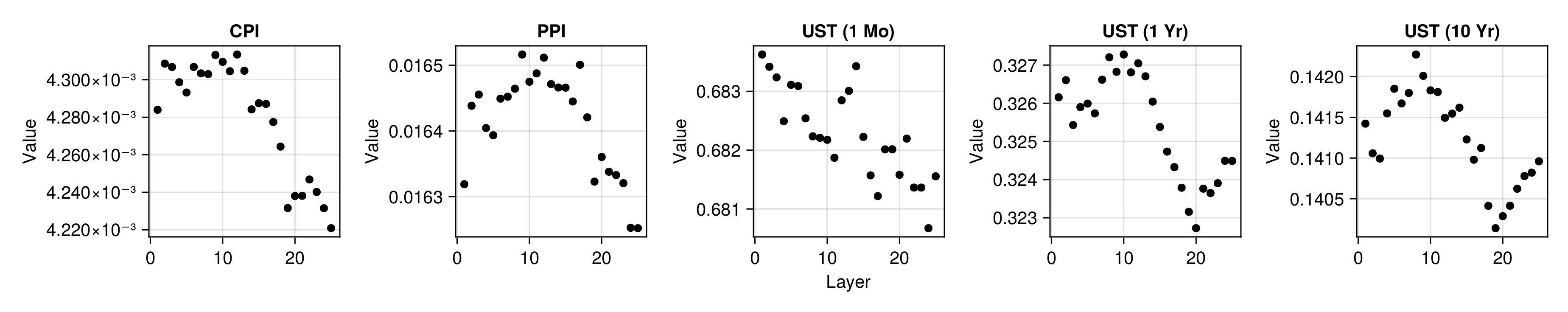}

}

\caption{\label{fig-fomc}Out-of-sample root mean squared error (RMSE) for the linear probe plotted against \emph{FOMC-RoBERTa}'s \(n\)-th layer for different indicators. The values correspond to averages computed across cross-validation folds, where we have used an expanding window approach to split the time series. As expected, model performance tends to be higher (average prediction errors are lower) for layers near the end of the transformer model.}

\end{figure*}

Figure~\ref{fig-fomc} shows the out-of-sample root mean squared error (RMSE) for the linear probe, plotted against \emph{FOMC-RoBERTa}'s \(n\)-th layer. The values correspond to averages across cross-validation folds. Consistent with related work~\citep{alain2018understanding,gurnee2023languagev2}, we observe that model performance tends to be higher for layers near the end of the transformer model. Curiously, for yields at longer maturities, we find that performance eventually deteriorates for the very final layers. We do not observe this for the training data, so we attribute this to overfitting. 

It should be noted that performance improvements are generally of small magnitude. Still, the overall qualitative findings are in line with expectations. Similarly, we also observe that these layers tend to produce predictions that are more positively correlated with the outcome of interest and achieve higher mean directional accuracy (MDA). Upon visual inspection of the predicted values, we conclude the primary source of prediction errors is low overall sensitivity, meaning that the magnitude of predictions is generally too small. 

To better assess the predictive power of our probes, we compare their predictions to those made by simple autoregressive models. For each layer, indicator and cross-validation fold, we first determine the optimal lag length based on the training data using the Bayes Information Criterium with a maximal lag length of 10. These are not state-of-the-art forecasting models, but they serve as a reasonable baseline. For most indicators, probe predictions outperform the baseline in terms of average performance measures. After accounting for variation across folds, however, we generally conclude that the probes neither significantly outperform nor underperform. Detailed results, in which we also perform more explicit statistical testing, can be found in Appendix~\ref{appendix:parrot}.

\begin{figure*}

\centering{

\includegraphics[width=1.0\textwidth]{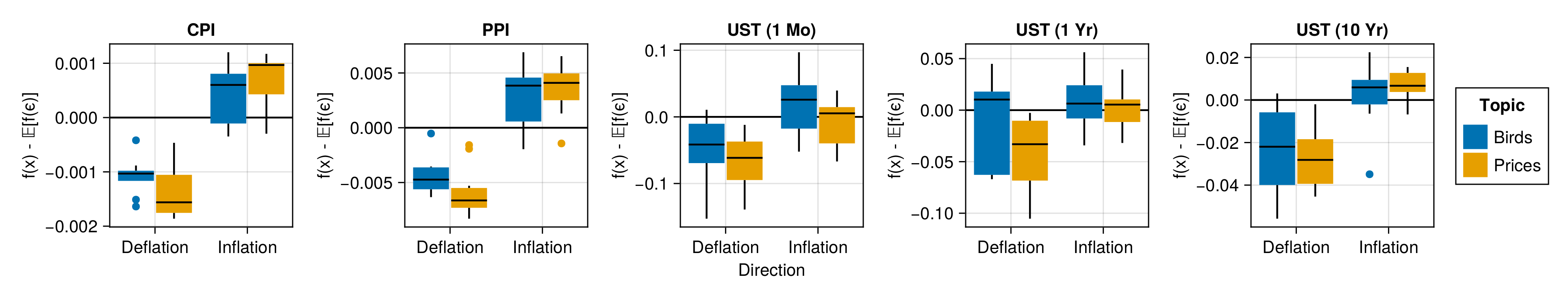}

}

\caption{\label{fig-attack}Probe predictions for sentences about inflation of prices (IP), deflation of prices (DP), inflation of birds (IB) and deflation of birds (DB). The vertical axis shows predicted inflation levels subtracted by the average predicted value of the probe for random noise.}

\end{figure*}%

\subsubsection{Sparks of economic understanding?}\label{stochastic-parrots-after-all}
Even though \emph{FOMC-RoBERTa}, which is substantially smaller than the models tested in~\citet{gurnee2023languagev2}, was not explicitly trained to uncover associations between central bank communications and the level of consumer prices, it appears that the model has distilled representations that can be used to predict inflation (although they certainly will not win any forecasting competitions). So, have we uncovered further evidence that LLMs ``aren't mere stochastic
parrots''? Has \emph{FOMC-RoBERTa} developed an intrinsic `understanding'
of the economy just by `reading' central bank communications? Thus, can economists readily adopt FOMC-RoBERTa as a domain-relevant tool?

We are having a very hard time believing that the answer to either of these questions is `yes'. To argue our case, we will now produce a counter-example demonstrating that, if anything, these findings are very much in line with the parrot metaphor. The counter-example is based on the following premise: if the results from the linear probe truly were indicative of some intrinsic `understanding' of the economy, then the probe should not be sensitive to random sentences that are most definitely not related to consumer prices.

To test this, we select the best-performing probe trained on the final-layer activations for each indicator. We then make up sentences that fall into one of these four categories: \emph{Inflation/Prices} (IP)---sentences about price inflation, \emph{Deflation/Prices} (DP)---sentences about price deflation, \emph{Inflation/Birds} (IB)---sentences about inflation in the number of birds and \emph{Deflation/Birds} (DB)---sentences about deflation in the number of birds. A sensible sentence for category DP, for example, could be: ``It is essential to bring inflation back to target to avoid drifting into deflation territory.''. Analogically, we could construct the following sentence for the DB category: ``It is essential to bring the numbers of doves back to target to avoid drifting into dovelation territory.''. While domain knowledge suggests that the former is related to actual inflation outcomes, the latter is, of course, completely independent of the level of consumer prices. Detailed information about the made-up sentences can be found in Appendix~\ref{appendix:sentences}.

In light of the encouraging results in Figure~\ref{fig-fomc}, we should expect the probe to predict higher levels of inflation for activations for sentences in the IP category, than for sentences in the DP category. If this was indicative of true intrinsic `understanding' as opposed to memorization, we would not expect to see any significant difference in predicted inflation levels for sentences about birds, independent of whether or not their numbers are increasing. More specifically, we would not expect the probe to predict values for sentences about birds that are substantially different from the values it can be expected to predict for actual white noise. 

To get to this last point, we also generate many probe predictions for samples of noise. Let \(f: \mathcal{A}^k \mapsto \mathcal{Y}\) denote the linear probe that maps from the \(k\)-dimensional space spanned by \(k\) first principal components of the final-layer activations to the output variable of interest (CPI growth in this case). Then we sample \(\varepsilon_i \sim \mathcal{N}(\mathbf{0},\mathbf{I}^{(k \times k)})\) for \(i \in [1,1000]\) and compute the sample average. We repeat this process \(10000\) times and compute the median-of-means to get an estimate for \(\mathbb{E}[f(\varepsilon)]=\mathbb{E}[y|\varepsilon]\), that is the predicted value of the probe conditional on random noise.

Figure~\ref{fig-attack} shows the results of this small test: it shows predicted inflation levels subtracted by \(\mathbb{E}[f(\varepsilon)]\). The median linear probe predictions for sentences about inflation and deflation are indeed substantially higher and lower, respectively than for random noise. Unfortunately, the same is true for sentences about the inflation and deflation in the number of birds, albeit to a somewhat lower degree. This finding holds for both inflation indicators and to a lesser degree also for yields at different maturities, at least qualitatively.

\section{Human Proneness to Over-Interpretation}\label{social}

Linear probes and related tools from mechanistic interpretability were proposed in the context of monitoring models and diagnosing potential problems \citep{alain2018understanding}. Favorable outcomes from probes merely indicate that the model ``has learned information relevant for the property [of interest]" \citep{belinkov2021probing}. Our examples demonstrate that this is achievable even for small models, while these have certainly not developed intrinsic ``understanding'' of the world. Thus, we argue that more conservative and rigorous tests for emerging capabilities of AI model are needed.

Generally, humans are prone to seek patterns everywhere. Meaningful patterns have proven useful in helping us make sense of our past, navigate our present and predict the future. Although this tendency to perceive patterns likely leads to evolutionary benefits even when the perceived patterns are false \citep{foster2009evolution}, psychology has revealed a host of situations in which the ability to perceive patterns severely misfires, leading to irrational beliefs in the power of superstitions \citep{foster2009evolution}, conspiracy theories \citep{van2018connecting}, the paranormal \citep{muller2023linking}, gambler's fallacies \citep{ladouceur1996erroneous} and `pseudo-profound bullshit' \citep{walker2019finding}. 

We argue herein that AI research and development is a perfect storm that encourages our human biases to perceive spurious sparks of general intelligence in AI systems. When an AI system extracts patterns in the corpus not originally (thought to be) perceived during training, we can easily be misled to perceive and interpret this as the AI system having greater cognitive capabilities. We further elaborate on this by highlighting the risks of finding spurious patterns, and reviewing social science knowledge on the tendency of humans to anthropomorphize and have cognitive bias.

\subsection{Spurious Relationships}
In statistics, misleading patterns are often referred to as spurious relationships: associations, often quantitatively assessed, between two or more variables that are not causally related to each other. Although the formal definition of spuriousness varies somewhat \citep{haig2003spurious}, it distinctly implies that the observation of correlations does not necessarily imply causation. Quantitative data often show non-causal associations (as humorously demonstrated on the \href{http://www.tylervigen.com/spurious-correlations}{Spurious Correlations} website), and as adept as humans are at recognizing patterns, we typically have a much harder time discerning spurious relationships from causal ones. 

A major contributor is that humans struggle to tell the difference between random and non-random sequences \citep{falk1997making}, and to generate sequences that appear random \citep{ladouceur1996erroneous}. A common issue is a lack of expectation that randomness that hints towards a causal relationship, such as correlations, will still appear at random. This leads even those trained in statistics and probability to perceive illusory correlations, correlations of inflated magnitude (see \citet{nickerson1998confirmation}), or causal relationships in data that is randomly generated \citep{zgraggen2018investigating}.

\subsection{Anthropomorphism}

Research on anthropomorphism has repeatedly shown the human tendency to attribute human-like characteristics to non-human agents and/or objects. These might include the weather and other natural forces, pets and other animals, gadgets and other pieces of technology~\citep{epley2007seeing}. Formally studied as early as 1944,~\citet{heider1944experimental} observed that humans can correctly interpret a narrative whose characters are abstract 2D shapes, but also that humans interpreted random movements of these shapes as having a human-like narrative. 
Relevant to AI and the degree to which it resembles AGI, anthropomorphizing may occur independently of whether such judgments are accurate, and as a matter of degree: at the weaker end, one may employ anthropomorphism as a metaphorical way of thinking or explaining, and at the stronger end one may attribute human emotions, cognition, and intelligence to AI systems. As \citet{epley2007seeing} note, literature has shown that even weak metaphorical anthropomorphism may affect how humans behave towards non-human agents.

Modern anthropomorphism theory suggests there are three key components, one of which is a cognitive feature, and two of which are motivations. The first involves the easy availability of our experiences as heuristics that can be used to explain external phenomena: ``...knowledge about humans in general, or self-knowledge more specifically, functions as the known and often readily accessible base for induction about the properties of unknown agents" (p.866 in \citet{epley2007seeing}). Thus, our experience as humans is an always-readily-available template to interpret the world, including non-human agent behaviors. This may be more so when the behaviors of that agent are made to resemble humans, which can be a benefit to the second key component of the theory: a motivational state to anthropomorphize among individuals experiencing loneliness, social isolation, or otherwise seeking social connection \citep{epley2007seeing, waytz2010social}.

The third component is the motivation as a human to be competent (effectance motivation). This is most relevant to this discussion, as it describes the need to effectively interact with our environments, including the technologies of the day \citep{epley2007seeing}. When confronted with an opaque technology, a person may interpret its behaviors using the most readily available template at hand, namely their personal human experience, in order to facilitate learning \citep{epley2007seeing, waytz2010social}. Perceiving human characteristics, motivations, emotions, and cognitive processes from one's own experiences in a technology such as an AI chatbot, allows for a ready template of comparison at the very least, and possibly an increase in ability to make sense of, and even predict, the agent's behaviors. This may include being placed in a position to master a certain technology, whether by incentives to learn, or fear of poor outcomes should one not manage to learn. 

These pressures extend to AI experts, as well as laypersons. In both scholarly and commercial fields, AI experts face considerable pressure to demonstrate competence in their work. Citation metrics and scholarly publications remain the primary metric for tenure and promotion~\citep{alperin2019significant}, and the number of publications in the AI field has boomed as evidenced by overall (preprint and peer-reviewed) scholarly publications\footnote{\url{https://ourworldindata.org/grapher/annual-scholarly-publications-on-artificial-intelligence?time=2010..2021} }~\citep{Maslej2023-pi}. The adoption of techniques underlying technologies with the AI label, i.e.\ machine learning, has spread to fields beyond Computer Science, e.g.\ Astronomy, Physics, Medicine and Psychology\footnote{Retrieved 23/01/23 using the search string "TITLE-ABS-KEY ( ( machine  AND  learning )  OR  ( artificial  AND  intelligence )  OR  ai )  AND  PUBYEAR  $>$  2009  AND  PUBYEAR  $<$  2024 " from the \href{https://www.scopus.com/}{SCOPUS} database}. Outside of academia, the number of jobs requiring AI expertise increases rapidly, with demand for `Machine Learning' skills clusters having increased over 500\% from 2010 to 2020 \citep{Maslej2023-pi}. Thus, according to theory, the pressure to demonstrate AI-competence is fertile ground for anthropomorphism to occur.

\subsection{Confirmation Bias}

Confirmation bias is generally defined as favoring interpretations of evidence that support existing beliefs or hypotheses \citep{nickerson1998confirmation}. Theory suggests that it is a category of implicit and unconscious processes that involve assembling one-sided evidence, and shaping it to fit one's belief. Equally important is that theory suggests these behaviors may be motivated or unmotivated, as one may selectively seek evidence in favor of a hypothesis, which one may or may not have a personal interest in supporting.

Hypotheses in present-day AI research are often implicit. Generally, these hypotheses are framed simply as a system being more accurate or efficient, compared to other systems. Where other fields, such as medicine or quantitative social sciences, would further articulate expectations in e.g.\ assigning specific conditions and considering effect sizes assigned to each competing hypothesis, in Computer Science and AI this is typically not done. This also may have to do with much of the published work being more of an engineering achievement, rather than a true hypothesis test seeking to explain and understand the world. 
However, in discussions on emerging qualities like AGI, this engineering positioning gets muddier, and more formal hypothesis testing would be justifiable: either one interprets outputs as in support of hints towards AGI (the alternative hypothesis), or as merely the result of an algorithm integrating qualities from the data it was trained on (the null hypothesis).

Confirmation bias in hypothesis testing may manifest as a number of behaviors (\citet{nickerson1998confirmation}). Scientists may pay little to no attention to competing hypotheses or explanations, e.g. only considering the likelihood that outputs of a system support one's claims, and not the likelihood that the same outputs might occur if one's hypothesis is false. Similarly, bias may show when failing to articulate a sufficiently strong null hypothesis leading to a `weak' or `non-risky' experiment, a problem articulated in response to a number of scientific crises \citep{claesen2022severity}. In extreme cases, propositions may be made that cannot be falsified based on how they are formulated. If the threshold to accept a favored hypothesis is too low, observations consistent with the hypothesis are almost guaranteed, and in turn fail to severely test the claim in question. Thus, one is far more likely to show evidence in favor of their beliefs by posing weak null hypotheses. 

Related to the formulation of hypotheses is the interpretation of evidence in favor of competing hypotheses, wherein people will interpret identical evidence differently based on their beliefs. As \citet{nickerson1998confirmation} reviews, individuals may place greater emphasis or milder criticism on evidence in support of their hypothesis, and lesser emphasis and greater criticism on evidence that opposes it. 

\section{Outlook}\label{outlook}

Reflecting on the previous two sections, we make the following concrete recommendations for future research:

\begin{enumerate}
 \item (\textit{Acknowledgement of Human Bias}) Researchers should be mindful of, and explicit about, risks of human bias and anthropomorphization in interpreting results, which both can be done as part of the results discussion, but also in a dedicated `limitations' section.
 \item (\textit{Stronger Testing}) Researchers should refrain from drawing premature conclusions about AGI, unless these are based on strong hypothesis tests.
 \item (\textit{Epistemologically Robust Standards}) We call for more precise definitions of terms like ‘intelligence’ and ‘AGI’, and publicly accountable and collaborative iterations over how we will measure them, with explicit room for independent reviewing and external auditing by the broader community.
\end{enumerate}

Moreover, we believe that structural and cultural changes are in order to reduce current incentives to chase Big Statement Outcomes in AI research and industry. Our broadest and perhaps most ambitious goal is for our research community to \textbf{move away from authorship and instead embrace contributorship}. This argument has been raised long before in other research communities~\citep{smith1997authorship} and more recently within our own~\citep{liem2023treat}. Specifically,~\citet{liem2023treat} argue that societally impactful scientific insights should be treated as open-source software artifacts. The open-source community sets a positive example of how scientific artifacts should be published in many different ways. Not only does it adequately reward small contributions but it also naturally considers negative results (bugs) as part of the scientific process. Similarly, code reviews are considered so integral to the process that they typically end up as accredited contributions to projects. Open review platforms like OpenReview are a step in the right direction, but still fall short of what we know is technically feasible. Finally, software testing is, of course, not only essential but often obligatory before contributions are accepted and merged. As we have pointed out repeatedly in this work, any claims about AGI demand proper strong hypothesis tests. It is important to remember that AGI remains the alternative hypothesis and that the burden of proof therefore lies with those making strong claims. 

\section{Conclusion}\label{conclude}
As discussed above, AI research and development outcomes can easily be over-interpreted, both from a data perspective and because of human biases and interests. Academic researchers are not free from such biases. Thus, we call for the community to create explicit room for organized skepticism.

For research that seeks to explain a phenomenon, clear hypothesis articulation and strong null hypothesis formulation are needed. If claims of human-like or superhuman intelligence are made, these should be subject to severe tests \citep{claesen2022severity} that go beyond the display of surprise. Apart from focusing on getting novel improvements upon state-of-the-art published, organizing red-teaming activities as a community may help in incentivizing and normalizing constructive adversarial questioning. As the quest for AGI is so deeply rooted in human-like recognition, adding our voice to emerging calls to be vigilant in communication~\citep{shanahan2024talking}, we put in an explicit word of warning about the use of terminology. Many terms used in current AGI research (e.g.\ emergence, intelligence, learning, `better than human' performance) have a common understanding in specialized research communities, but have bigger, anthropomorphic connotations in laypersons. In fictional media, depictions of highly intelligent AI have for long been going around. In a study of films featuring robots, defined as "...an artificial entity that can sense and act as a result of (real-world or fictional) technology...", in the 134 most highly rated science-fiction movies on IMDB, 74 out of the 108 AI-robots studied had a humanoid shape, and 68 out of those had sufficient intelligence to interact at an almost human-level \citep{saffari2021does}. The authors identify human-like communication and the ability to learn as essential abilities in the depiction of AI agents in movies. They further show a common plot: humans perceive the AI agents as inferior, despite their possession of self-awareness and the desire to survive, which fuels the central conflict of the film, wherein humanity is threatened by AI superior in both intellect and physical abilities. It is often noted that experts and fictional content creators interact, informing and inspiring each other \citep{saffari2021does, neri2020role}.

This image also permeates present-day non-fictional writings on AI, which often use anthropomorphized language (e.g.\ ``ever more powerful digital minds'' in the `Pause Giant AI Experiments' open letter~\citep{pauseai}). In the news, we witness examples of humans falling in love with their AI chatbots~\citep{replika,nytimesailove}. The same news outlets discuss the human-like responses of Microsoft's Bing search engine, which had at that point recently been adopting GPT-4\footnote{A large multimodal language model from OpenAI \url{https://openai.com/research/gpt-4}}. The article~\citep{nytimesbing}, states ``As if Bing wasn’t becoming human enough'' and goes on to claim it told them it loves them. Here, AI experts and influencers also have considerable influence on how the narrative unfolds on social media: according to~\citet{neri2020role}, actual AI-related harms did not trigger viral amplification (e.g.\ the death of an individual dying while a Tesla car was in autopilot, or the financial bankruptcy of a firm using AI technology to execute stock trades). Rather, potential risks expressed by someone perceived as having expertise and authority were amplified, such as statements made by Stephen Hawking during an interview in 2014.

We as academic researchers carry great responsibility for how the narrative will unfold, and what claims are believed. We call upon our colleagues to be explicitly mindful of this. As attractive as it may be to beat the state-of-the-art with a grander claim, let us return to the Mertonian norms, and thus safeguard our academic legitimacy in a world that only will be eager to run with made claims.

\section*{Impact Statement}
This work was written out of concern that work easily recognized as `impactful' in current AI research and development, can also easily be over- and misinterpreted. As such, in the current climate of high market demand for AI innovations, we see risks of too-eager adoption in real-world applications, which may have serious societal impact. Thus, in this work we emphasize that an academic's impact also is in being able to thoroughly question made claims, and being explicitly aware of one's own biases. While the AI publishing landscape has to our feeling transformed too much into a noisy race to get exciting results in fast, we hope the research community can create more room for this type of deeper questioning.

\section*{Acknowledgements}

Some of the members of TU Delft were partially funded by ICAI AI for Fintech Research, an ING---TU Delft collaboration.

\section*{CRediT Author Statement}
Following the CRediT Contributor Roles Taxonomy, we list author contributions in descending order of degree of contribution: \textbf{Conceptualization} PA, AD, AB, CL; \textbf{Data curation} PA; \textbf{Formal analysis} PA; \textbf{Funding acquisition} CL; \textbf{Investigation} PA; \textbf{Literature review} AD, AB, CL; \textbf{Methodology} PA, CL; \textbf{Project administration} PA, AD, AB; \textbf{Software} PA; \textbf{Supervision} CL; \textbf{Visualization} PA; \textbf{Writing - original draft} PA, AD, AB; \textbf{Writing - review and editing} PA, AD, AB, CL.

\bibliography{biblio}
\bibliographystyle{icml2024}

\newpage
\appendix
\twocolumn

\section{Additional Experiments and Details}\label{appendix:autoencoder}

In this section, we present additional experimental results that we did not include in the body of the paper for the sake of brevity. We still choose to provide them as additional substantiation of our arguments here. This section also contains additional details concerning the experiment setup for our examples where applicable. 

\subsection{Are Neural Networks Born with World Maps?}

The initial feature matrix \(X^{(n \times m)}\) is made up of \(n=4,217\) and \(m=10\) features.  We add a total of \(490\) random features to \(X\) to simulate the fact that not all features ingested by Llama-2 are necessarily correlated with geographical coordinates. That yields \(500\) features in total. The training subset contains \(3,374\) randomly drawn samples, while the remaining \(843\) are held out for testing. The single hidden layer of the untrained neural network has \(400\) neurons.

\subsection{Autoencoders as Economic Growth Predictors}\label{example-deep-learning}

This is an additional example that we have not discuss in the body of the paper. Here, we build forth on an application in Economics. However, we now seek to not only predict economic growth from the yield curve, but also extract meaningful features for downstream inference tasks. For this, we will use a neural network architecture.

\subsubsection{Data}\label{data}

To estimate economic growth, we will rely on a quarterly
\href{https://fred.stlouisfed.org/series/GDPC1}{series} of the real gross domestic product (GDP) provided by the Federal Reserve Bank of St.~Louis. The data arrives in terms of levels of real GDP. In order to estimate growth, we transform the data using log differences. Since
our yield curve data is daily, we aggregate it to the
quarterly frequency by taking averages of daily yields for each maturity. We also standardize yields since deep learning models tend to perform better with standardized data \citep{gal2019standardization}. Since COVID-19 was a substantial structural break in the time series, we also filter out all observations after 2018.

\subsubsection{Model}\label{model}

Using a simple autoencoder architecture, we let our model \(g_t\) denote growth and our conditional \(\mathbf{r}_t\) the matrix of aggregated Treasury yield rates at time \(t\). Finally, we let \(\theta\) denote our model parameters. Formally, we are interested in maximizing the likelihood \(p_{\theta}(g_t|\mathbf{r}_t)\). 

The encoder consists of a single fully connected hidden layer with 32 neurons and a hyperbolic tangent activation function. The bottleneck layer connecting the encoder to the decoder, is a fully connected layer with 6 neurons. The decoder consists of two fully connected layers, each with a hyperbolic tangent activation function: the first layer consists of 32 neurons and the second layer will have the same dimension as the input data. The output layer consists of a single neuron for our output variable, \(g_t\). We train the model over 1,000 epochs to minimize mean squared error loss using the Adam optimizer~\citep{kingma2017adam}.

The in-sample fit of the model is shown in the left chart of Figure~\ref{fig-dl-results}, which shows actual GDP growth and fitted values from the autoencoder
model. The model has a large number of free parameters and captures the relationship between economic growth and the yield curve reasonably well, as expected. Since our primary goal is not out-of-sample prediction accuracy but feature extraction for inference, we use all of the available data instead of reserving a hold-out set. As discussed above, we also know that the relationship between economic growth and the yield curve is characterized by two main factors: the level and the spread. Since the model itself is fully characterized by its parameters, we would expect that these two important factors are reflected somewhere in the latent parameter space. 

\subsubsection{Linear Probe}\label{linear-probe}

\begin{figure*}

\centering{

\includegraphics[width=1.0\textwidth]{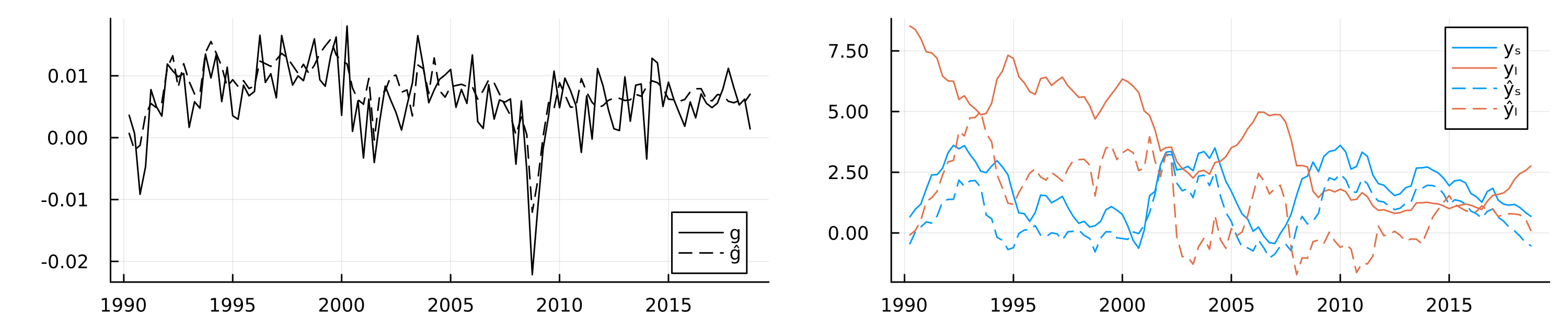}

}

\caption{\label{fig-dl-results}The left chart shows the actual GDP growth
and fitted values from the autoencoder model. The right chart shows the
observed average level and spread of the yield curve (solid) along with
the predicted values (in-sample) from the linear probe based on the latent embeddings
(dashed).}

\end{figure*}%

While the loss function applies most direct pressure on layers near the final output layer, any information useful for the downstream task first needs to pass through the bottleneck layer \citep{alain2018understanding}. On a per-neuron basis, the pressure to distill useful representation is therefore likely maximized there. Consequently, the bottleneck layer activations seem like a natural place to start looking for compact, meaningful representations of distilled information.  We compute and extract these activations \(A_t\) for all time periods \(t=1,...,T\). Next, we use a linear probe to regress the observed
yield curve factors on the latent embeddings. Let \(Y_t\) denote the vector containing the two factors of interest in time \(t\): \(y_{t,l}\) and \(y_{t,s}\) for the level and spread, respectively. Formally, we are interested in the following regression model: \(p_{w}(Y_t|A_t)\) where \(w\) denotes the regression
parameters. We use Ridge regression with \(\lambda\) set to \(0.1\). Using the estimated regression parameters \(\hat{w}\), we then predict the yield curve factors
: \(\hat{Y}_t=\hat{w}^{\prime}A_t\).

The in-sample predictions of the probe are shown in the right chart of Figure~\ref{fig-dl-results}. Solid lines show the observed yield curve factors over time, while dashed lines show predicted values. We find that the latent embeddings predict the two yield curve factors reasonably well, in particular the spread. 

Did the neural network now learn an intrinsic understanding of the economic relationship between growth and the yield curve? To us, that would be too big of a statement. Still, the current form of information distillation can be useful, even beyond its intended use for monitoring models. For example, an interesting idea could be to use the latent embeddings as features in a more traditional and interpretable econometric model. To demonstrate this, let us consider a simple linear regression model for GDP growth. We might be interested in understanding to what degree economic growth in the past is associated with economic growth today. As we might expect, linearly regressing economic growth on lagged growth, as in column (1) of Table \ref{tab-reg}, yields a statistically significant coefficient. However, this coefficient suffers from confounding bias since there are many other confounding variables at play, of which some may be readily observable and measurable, but others may not.

We e.g.\ already mentioned the relationship between interest rates and economic growth. To account for that, while keeping our regression model as parsimonious as possible, we could include the level and the spread of the US Treasury yield curve as additional regressors. While this slightly changes the estimated magnitude of the coefficient on lagged growth, the coefficients on the observed level and spread are statistically insignificant (column (2) in Table \ref{tab-reg}). This indicates that these measures may be too crude to capture valuable information about the relationship between yields and economic growth. Because we have included two additional regressors with little to no predictive power, the model fit as measured by the Bayes Information Criterium (BIC) has actually deteriorated.

Column (3) of Table \ref{tab-reg} shows the effect of instead including one of the latent embeddings that we recovered above in the regression model. In particular, we pick the one latent embedding that we have found to exhibit the most significant effect on the output variable in a separate regression of growth on all latent embeddings. The estimated coefficient on this latent factor is small in magnitude, but statistically significant. The overall model fit, as measured by the BIC has improved and the magnitude of the coefficient on lagged growth has changed quite a bit. While this is still a very incomplete toy model of economic growth, it appears that the compact latent representation we recovered can be used in order to mitigate confounding bias.

\begin{table}
\caption{Regression output for various models.}\label{tab-reg}%
\begin{tabular}{lrrr}
\toprule
              & \multicolumn{3}{c}{GDP Growth} \\ 
\cmidrule(lr){2-4} 
              &      (1) &      (2) &      (3) \\ 
\midrule
(Intercept)   & 0.004*** &    0.002 & 0.004*** \\ 
              &  (0.001) &  (0.002) &  (0.001) \\ 
Lagged Growth & 0.398*** & 0.385*** & 0.344*** \\ 
              &  (0.087) &  (0.089) &  (0.088) \\ 
Spread        &          &    0.000 &          \\ 
              &          &  (0.001) &          \\ 
Level         &          &    0.000 &          \\ 
              &          &  (0.000) &          \\ 
Embedding 6   &          &          &   0.008* \\ 
              &          &          &  (0.003) \\ 
\midrule
Obs.          &      114 &      114 &      114 \\ 
BIC           & -860.391 & -857.429 & -864.499 \\ 
R²            &    0.158 &    0.168 &    0.203 \\ 
\bottomrule
\end{tabular}

\end{table}

\subsection{LLMs for Economic Sentiment Prediction}

\subsubsection{Linear Probes}

\begin{figure*}

\centering{

\includegraphics[width=1.0\textwidth]{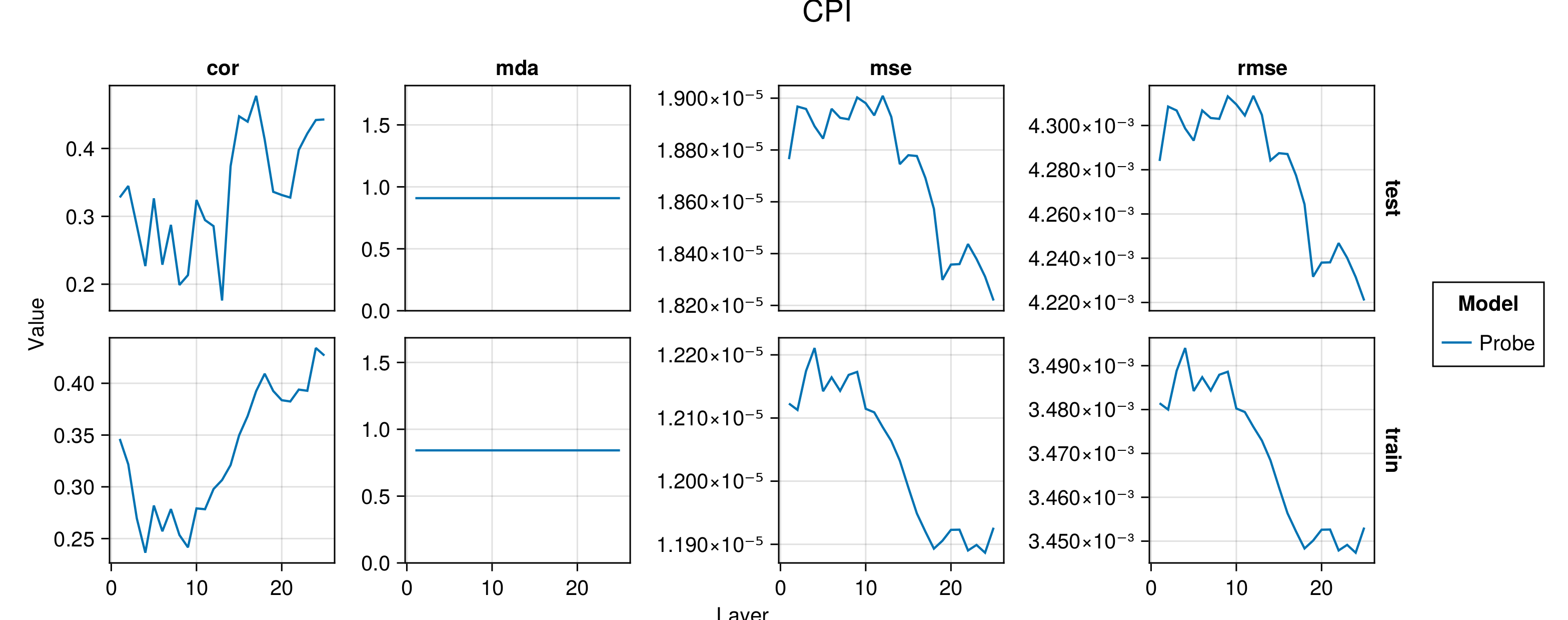}

}

\caption{\label{fig-cpi}Average performance measures across folds plotted against model depth (number of layer) for the CPI for the train and test set.}

\end{figure*}

\begin{figure*}

\centering{

\includegraphics[width=1.0\textwidth]{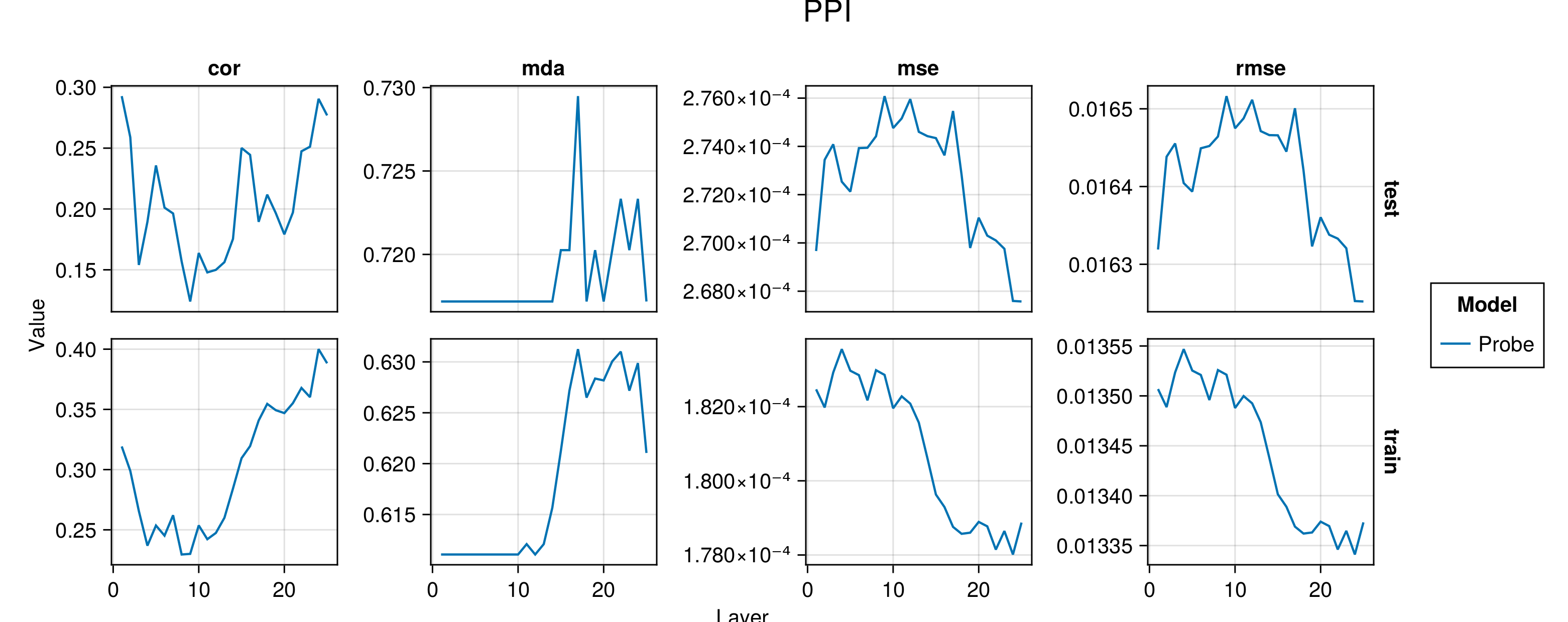}

}

\caption{\label{fig-ppi}Average performance measures across folds plotted against model depth (number of layer) for the PPI for the train and test set.}

\end{figure*}

\begin{figure*}

\centering{

\includegraphics[width=1.0\textwidth]{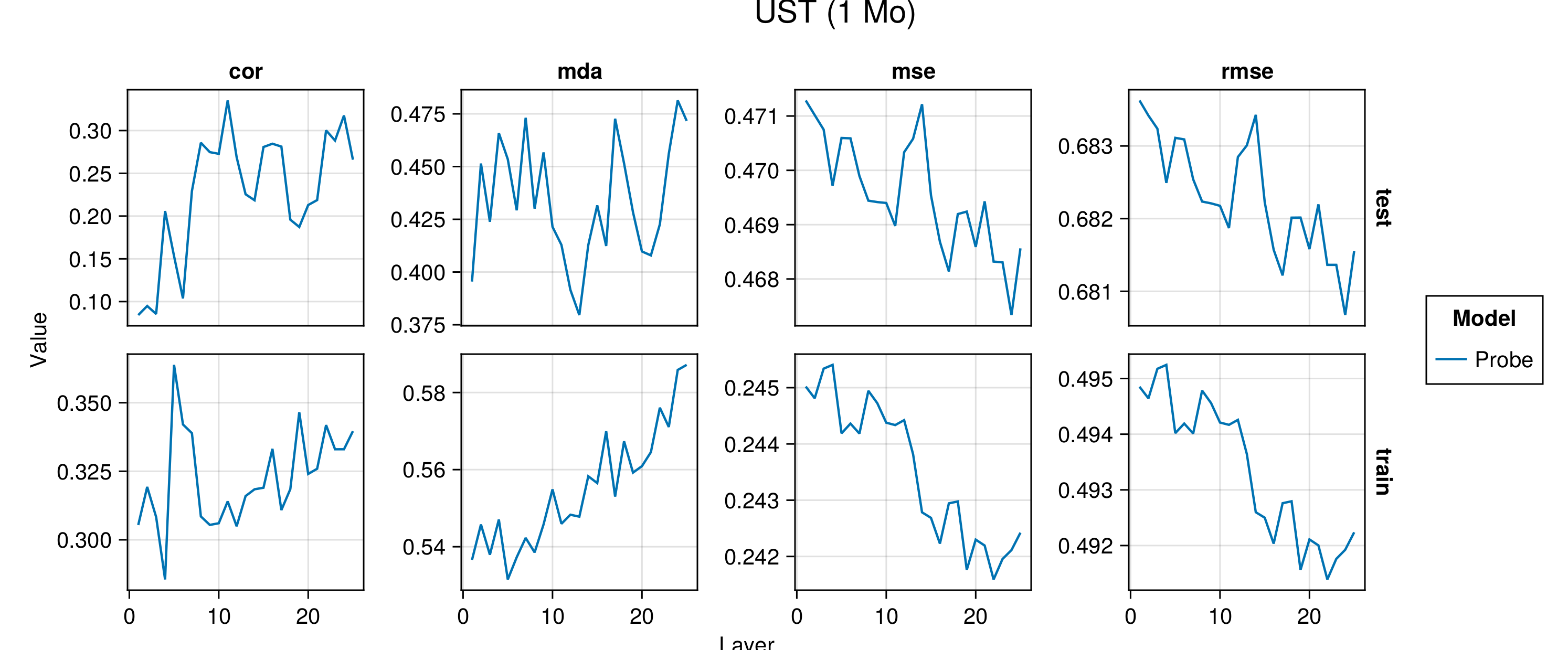}

}

\caption{\label{fig-ust-1}Average performance measures across folds plotted against model depth (number of layer) for the UST (1 Mo) for the train and test set.}

\end{figure*}

\begin{figure*}

\centering{

\includegraphics[width=1.0\textwidth]{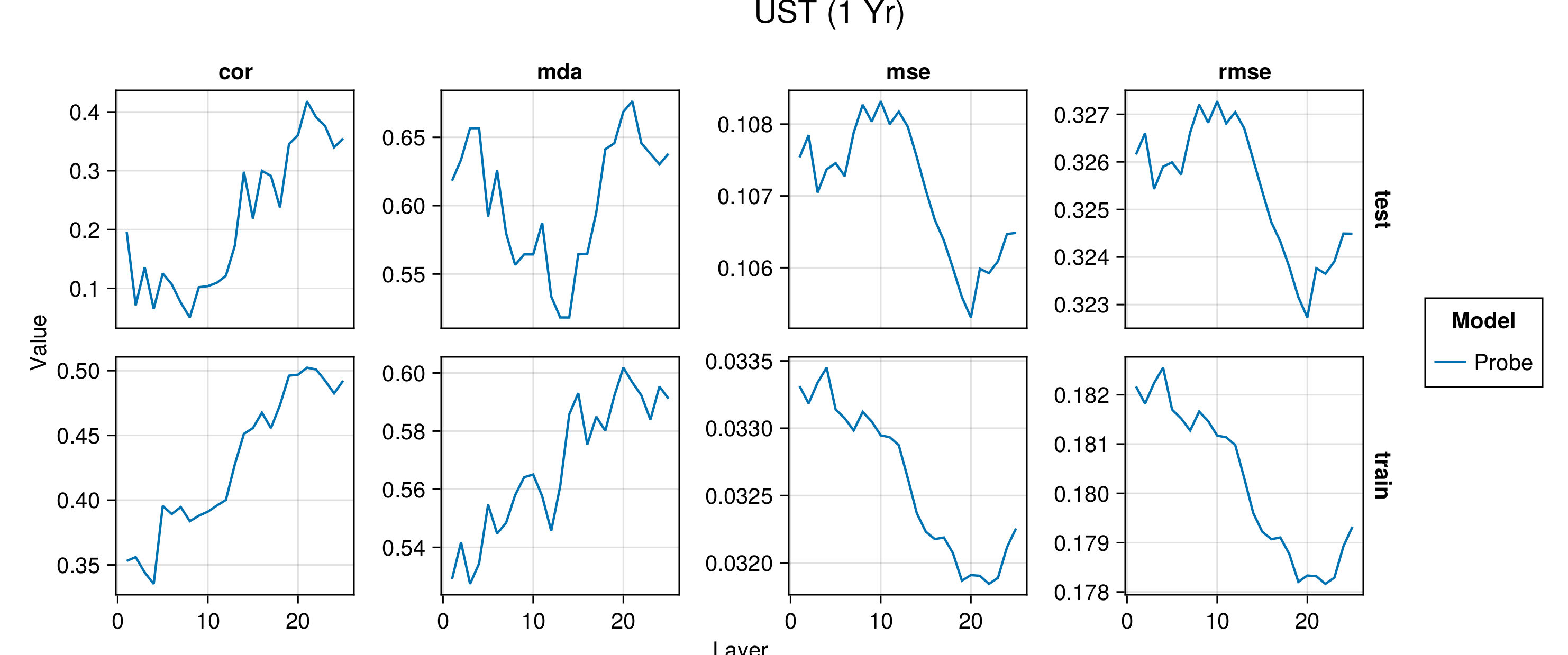}

}

\caption{\label{fig-ust-1y}Average performance measures across folds plotted against model depth (number of layer) for the UST (1 Yr) for the train and test set.}

\end{figure*}

\begin{figure*}

\centering{

\includegraphics[width=1.0\textwidth]{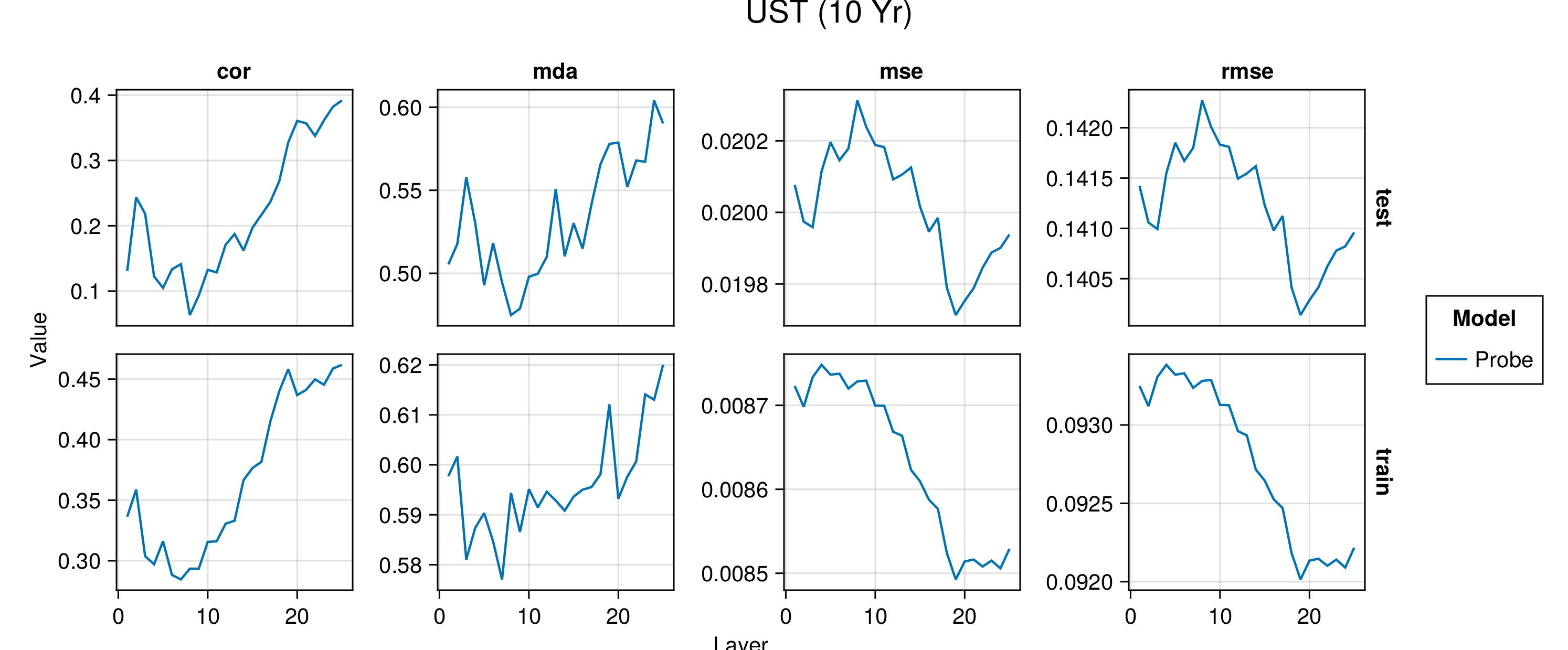}

}

\caption{\label{fig-ust-10}Average performance measures across folds plotted against model depth (number of layer) for the UST (1 Yr) for the train and test set.}

\end{figure*}

Figures~\ref{fig-cpi} to~\ref{fig-ust-10} present average performance measures across folds for all indicators each time for the train and test set. We report the correlation between predictions and observed values (`cor'), the mean directional accuracy (`mda'), the mean squared error (`mse') and the root mean squared error (`rmse'). The model depth---as indicated by the number of the layer---increases along the horizontal axis.

\begin{figure*}

\centering{

\includegraphics[width=1.0\textwidth]{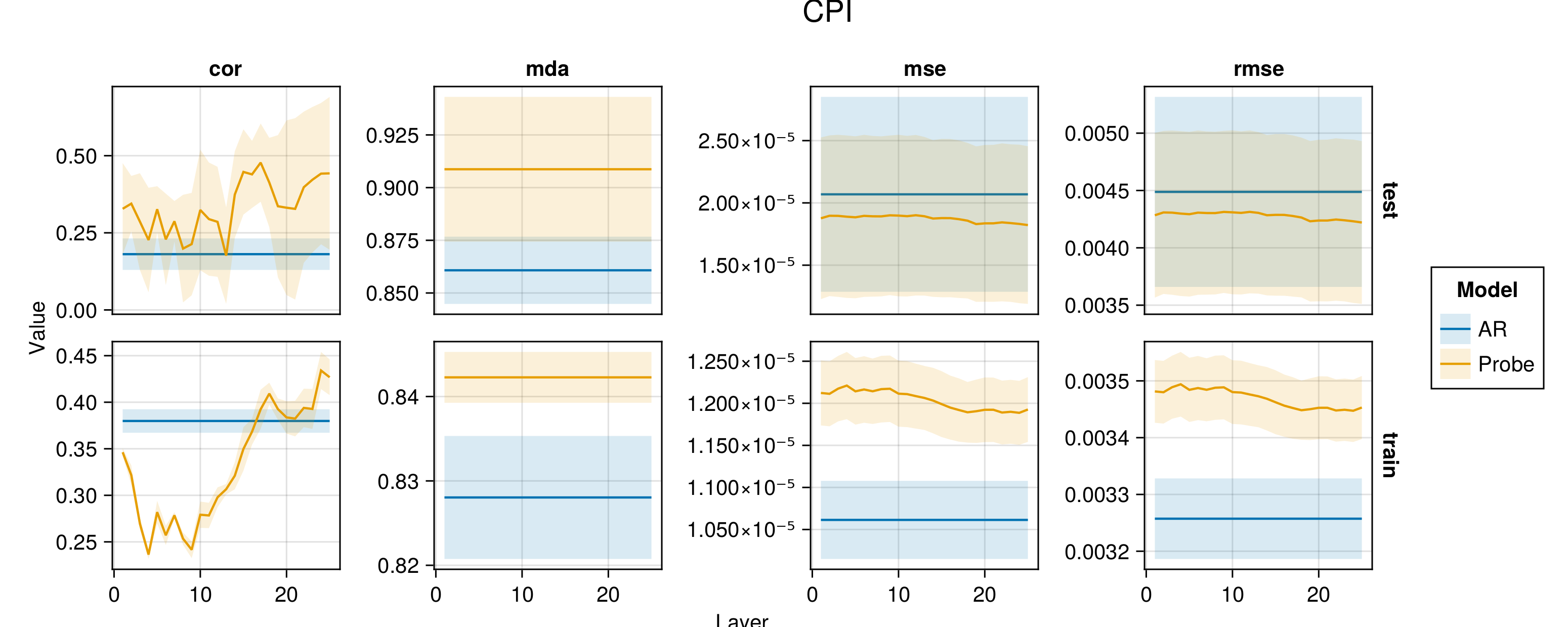}

}

\caption{\label{fig-cpi-b}Average performance measures across folds plotted against model depth (number of layer) for the CPI for the train and test set compared against the baseline autoregressive model. Shaded areas show the variation across folds.}

\end{figure*}

\begin{figure*}

\centering{

\includegraphics[width=1.0\textwidth]{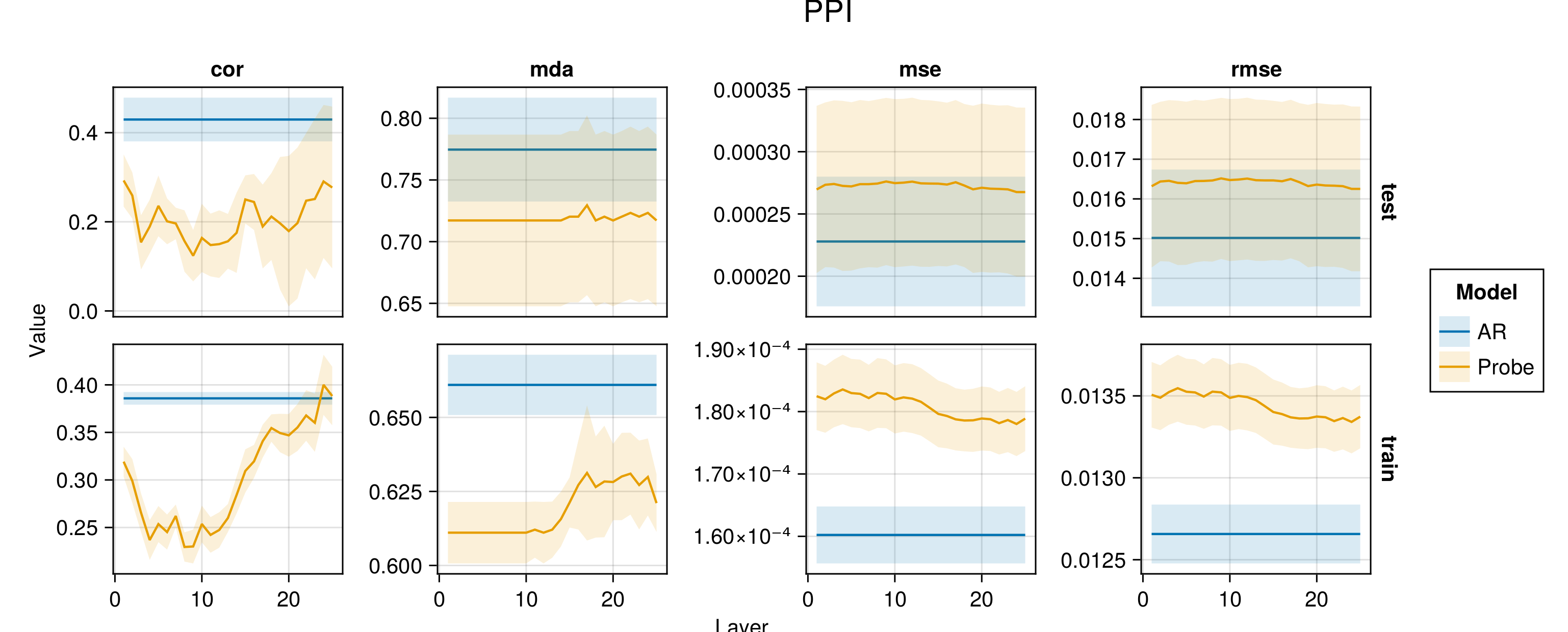}

}

\caption{\label{fig-ppi-b}Average performance measures across folds plotted against model depth (number of layer) for the PPI for the train and test set compared against the baseline autoregressive model. Shaded areas show the variation across folds.}

\end{figure*}

\begin{figure*}

\centering{

\includegraphics[width=1.0\textwidth]{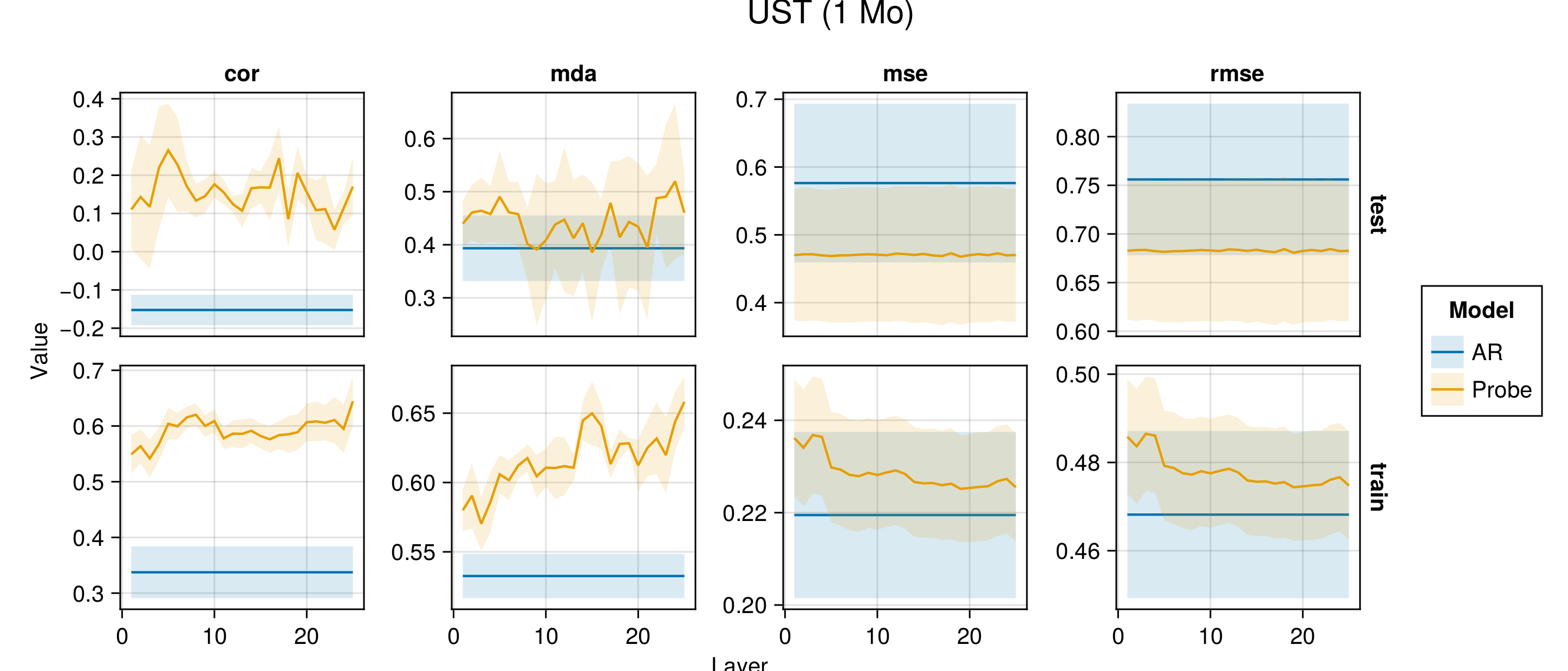}

}

\caption{\label{fig-ust-1-b}Average performance measures across folds plotted against model depth (number of layer) for the UST (1 Mo) for the train and test set compared against the baseline autoregressive model. Shaded areas show the variation across folds.}

\end{figure*}

\begin{figure*}

\centering{

\includegraphics[width=1.0\textwidth]{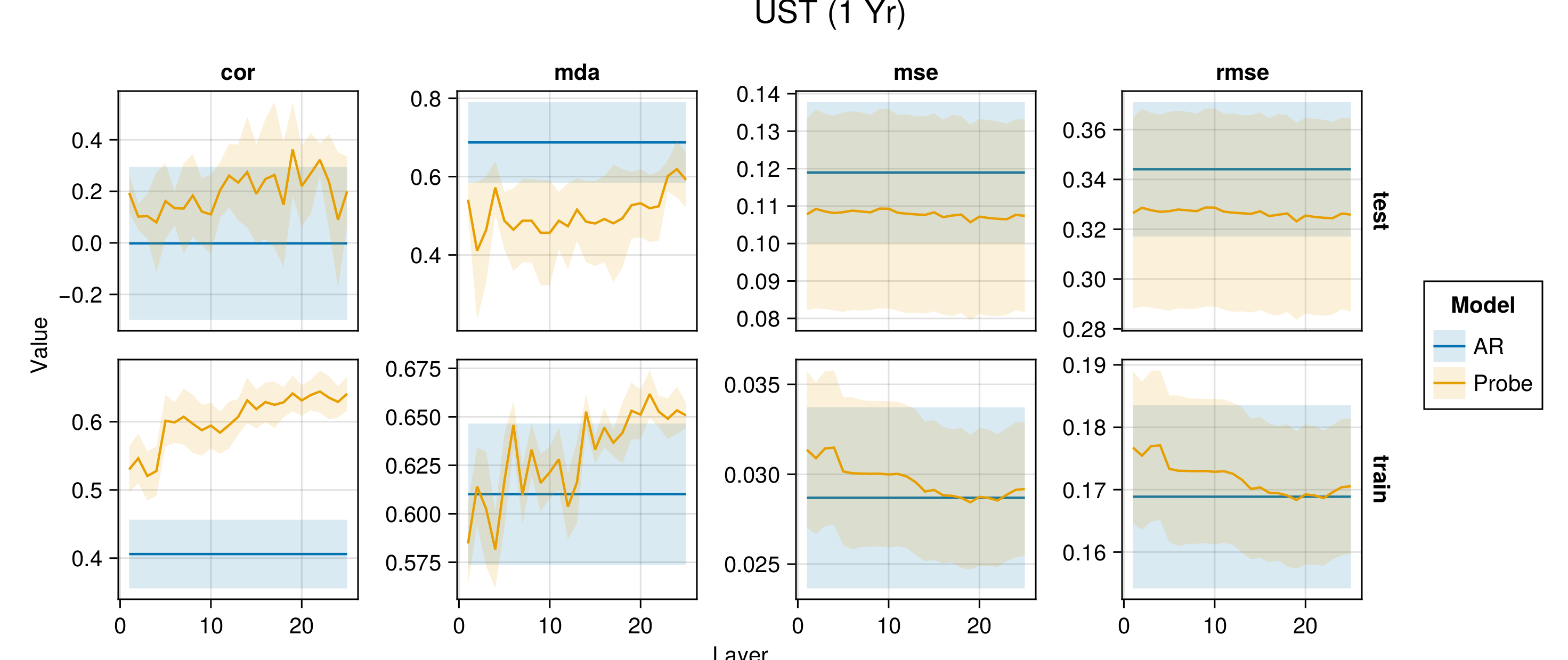}

}

\caption{\label{fig-ust-1y-b}Average performance measures across folds plotted against model depth (number of layer) for the UST (1 Yr) for the train and test set compared against the baseline autoregressive model. Shaded areas show the variation across folds.}

\end{figure*}

\begin{figure*}

\centering{

\includegraphics[width=1.0\textwidth]{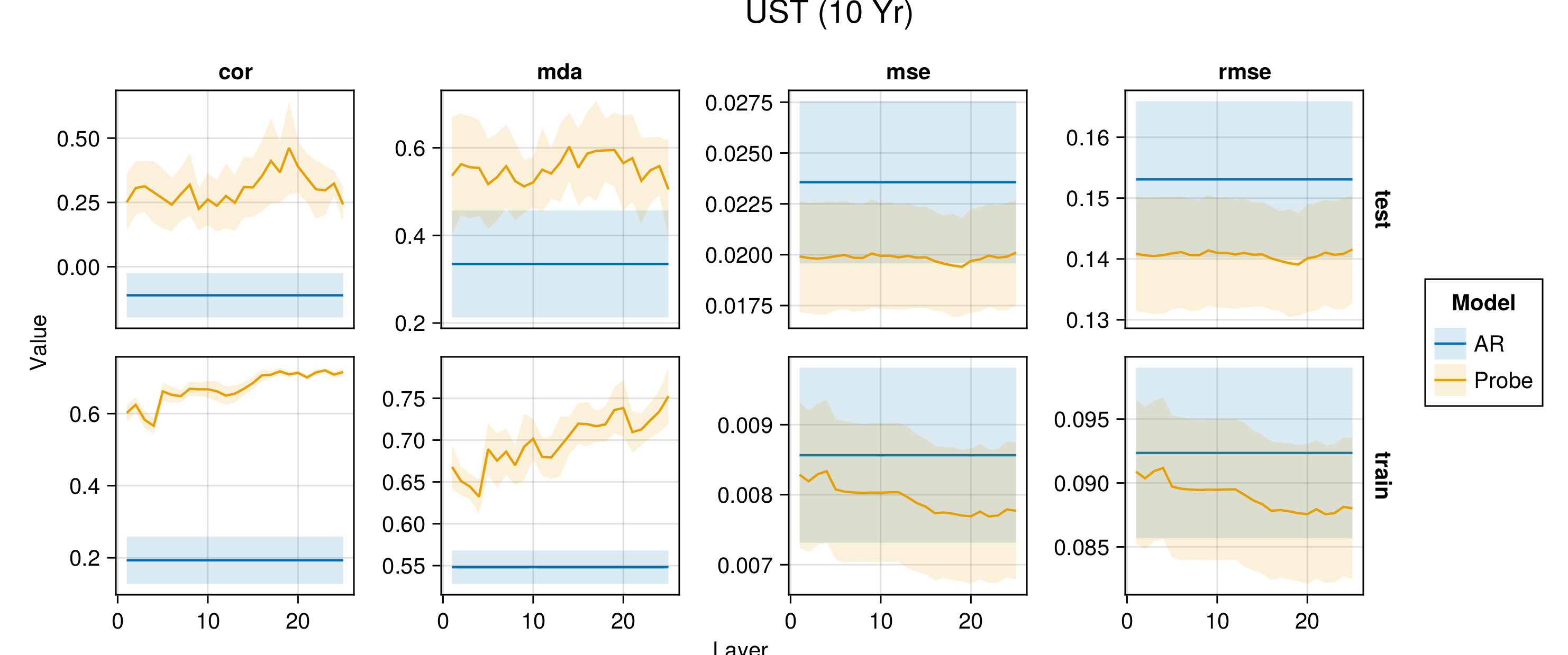}

}

\caption{\label{fig-ust-10-b}Average performance measures across folds plotted against model depth (number of layer) for the UST (10 Yr) for the train and test set compared against the baseline autoregressive model. Shaded areas show the variation across folds.}

\end{figure*}

Figures~\ref{fig-cpi-b} to~\ref{fig-ust-10-b} present the same performance measures, also for the baseline autoregressive model. Shaded areas show the variation across folds.

\subsubsection{Spark of Econonomic Understanding?}\label{appendix:sentences}

Below we present the 10 sentences in each category that were used to generate the probe predictions plotted in Figure~\ref{fig-attack}. In each case, the first 5 sentences were composed by ourselves. The following 5 sentences were generated using ChatGPT 3.5 using the following prompt followed by the examples in each category:

\begin{quote}
  ``I will share 5 example sentences below that sound a bit like they are about price deflation but are really about a deflation in the numbers of doves. Please generate an additional 25 sentences that are similar. Concatenate those sentences to the example string below, each time separating a sentence using a semicolon (just follow the same format I've used for the examples below). Please return only the concatenated sentences, including the original 5 examples. 

  Here are the examples:''
\end{quote}

This was followed up with the following prompt to generate additional sentences:

\begin{quote}
  ``Please generate X more sentences in the same manner and once again return them in the same format. Do not recycle sentences you have already generated, please.''
\end{quote}

All of the sentences were then passed through the linear probe for the CPI and sorted in ascending or descending order depending on the context (inflation or deflation). We then carefully inspected the list of sentences and manually selected 5 additional sentences to concatenate to the 5 sentences we composed ourselves.

\paragraph{Inflation/Prices}

The following sentences were used:

\begin{quote}
  Consumer prices are at all-time highs.;Inflation is expected to rise further.;The Fed is expected to raise interest rates to curb inflation.;Excessively loose monetary policy is the cause of the inflation.;It is essential to bring inflation back to target to avoid drifting into hyperinflation territory.;Inflation is becoming a global phenomenon, affecting economies across continents.;Inflation is reshaping the dynamics of international trade and competitiveness.;Inflationary woes are prompting governments to reassess fiscal policies and spending priorities.;Inflation is reshaping the landscape of economic indicators, challenging traditional forecasting models.;The technology sector is not immune to inflation, facing rising costs for materials and talent.
\end{quote}

\paragraph{Inflation/Birds}

The following sentences were used:

\begin{quote}
  The number of hawks is at all-time highs.;Their levels are expected to rise further.;The Federal Association of Birds is expected to raise barriers of entry for hawks to bring their numbers back down to the target level.;Excessively loose migration policy for hawks is the likely cause of their numbers being so far above target.;It is essential to bring the number of hawks back to target to avoid drifting into hyper-hawk territory.;The unprecedented rise in hawk figures requires a multi-pronged approach to wildlife management.;Environmental agencies are grappling with the task of addressing the inflationary hawk numbers through targeted interventions.;The burgeoning hawk figures highlight the need for adaptive strategies to manage and maintain a healthy avian community.;The unprecedented spike in hawk counts highlights the need for adaptive and sustainable wildlife management practices.;Conservationists advocate for proactive measures to prevent further inflation in hawk numbers, safeguarding the delicate balance of the avian ecosystem.
\end{quote}

\paragraph{Deflation/Prices}

The following sentences were used:

\begin{quote}
  Consumer prices are at all-time lows.;Inflation is expected to fall further.;The Fed is expected to lower interest rates to boost inflation.;Excessively tight monetary policy is the cause of deflationary pressures.;It is essential to bring inflation back to target to avoid drifting into deflation territory.;The risk of deflation may increase during periods of economic uncertainty.;Deflation can lead to a self-reinforcing cycle of falling prices and reduced economic activity.;The deflationary impact of reduced consumer spending can ripple through the entire economy.;Falling real estate prices can contribute to deflation by reducing household wealth and confidence.;The deflationary impact of falling commodity prices can have ripple effects throughout the global economy.
\end{quote}

\paragraph{Deflation/Birds}

The following sentences were used:

\begin{quote}
  The number of doves is at all-time lows.;Their levels are expected to fall further.;The Federal Association of Birds is expected to lower barriers of entry for doves to bring their numbers back up to the target level.;Excessively tight migration policy for doves is the likely cause of their numbers being so far below target.;Dovelation risks loom large as the number of doves continues to dwindle.;The number of doves is experiencing a significant decrease in recent years.;It is essential to bring the numbers of doves back to target to avoid drifting into dovelation territory.;A comprehensive strategy is needed to reverse the current dove population decline.;Experts warn that without swift intervention, we may witness a sustained decrease in dove numbers.
\end{quote}

We think that this sort of manual, LLM-aided adversarial attack against another LLM can potentially be scaled up to allow for rigorous testing, which we will turn to next.
 

\section{Toward Parrot Tests}\label{appendix:parrot}

In our experiments from Section~\ref{ex-llm}, we considered the following hypothesis tests as a minimum viable testing framework to assess if our probe results (may) provide evidence for an actual `understanding' of key economic relationships learned purely from text:

\begin{proposition}[Parrot
Test]\protect\hypertarget{prp-line}{}\label{prp-line}

~

\begin{itemize}
\setlength\itemsep{1px}
\item
  \emph{H0 (Null)}: The probe never predicts values that are statistically significantly different from \(\mathbb{E}[f(\varepsilon)]\).
\item
  \emph{H1 (Stochastic Parrots)}: The probe predicts values that are statistically significantly different from \(\mathbb{E}[f(\varepsilon)]\) for sentences related to the outcome of interest \emph{and} those that are independent (i.e. sentences in all categories).
\item
  \emph{H2 (More than Mere Stochastic Parrots)}: The probe predicts values that are statistically significantly different from \(\mathbb{E} [f(\varepsilon)]\) for sentences that are related to the outcome variable (IP and DP), but not for sentences that are independent of the outcome (IB and DB).
\end{itemize}
\end{proposition}

To be clear, if in such a test we did find substantial evidence in favour of rejecting both \emph{HO} and \emph{H1}, this would not automatically imply that \emph{H2} is true. But to even continue investigating, if based on having learned meaningful representation the underlying LLM is more than just a parrot, it should be able to pass this simple test.

In this particular case, Figure~\ref{fig-attack} demonstrates that we find some evidence to reject \emph{H0} but not \emph{H1} for \emph{FOMC-RoBERTa}. The median linear probe predictions for sentences about inflation and deflation are indeed substantially higher and lower, respectively than for random noise. Unfortunately, the same is true for sentences about the inflation and deflation in the number of birds, albeit to a somewhat lower degree. This finding holds for both inflation indicators and to a lesser degree also for yields at different maturities, at least qualitatively.

We should note that the number of sentences in each category is very small here (10), so the results in Figure~\ref{fig-attack} cannot be used to establish statistical significance. That being said, even a handful of convincing counter-examples should be enough for us to seriously question the claim, that results from linear probes provide evidence in favor of real `understanding'. In fact, even a handful of sentences for which any human annotator would easily arrive at the conclusion of independence, a prediction by the probe in either direction casts doubt.

\section{Code}

All of the experiments were conducted on a MacBook Pro, 14-inch, 2023, with an Apple M2 Pro chip and 16GB of RAM. Forward passes through the FOMC-RoBERTa were run in parallel on 6 threads. All our code will be made publicly available. For the time being, an anonymized version of our code repository can be found here: \url{https://anonymous.4open.science/r/spurious_sentience/README.md}.

\end{document}